\documentclass[dvipsnames]{article} 
\usepackage{iclr2023_conference_tinypaper,times}

\usepackage{amsmath,amsfonts,bm}

\def\eqref#1{equation~\ref{#1}}

\def\1{\bm{1}}

\DeclareMathAlphabet{\mathsfit}{\encodingdefault}{\sfdefault}{m}{sl}
\SetMathAlphabet{\mathsfit}{bold}{\encodingdefault}{\sfdefault}{bx}{n}

\usepackage{hyperref}
\usepackage{url}
\usepackage{xcolor}
\usepackage{graphicx}
\usepackage{float}
\usepackage{wrapfig}
\usepackage{booktabs}

\title{Large Language Models Perform\\Diagnostic Reasoning}

\author{Cheng-Kuang Wu\thanks{Equal contribution, authors listed alphabetically.},\,\,\,\,Wei-Lin Chen\footnotemark[1],\,\,\,\,Hsin-Hsi Chen \\
National Taiwan University, Taiwan\\
\texttt{\{ckwu,wlchen\}@nlg.csie.ntu.edu.tw, hhchen@ntu.edu.tw} 
}

\iclrfinalcopy 
\begin{document}

\maketitle

\begin{abstract}
We explore the extension of \textit{chain-of-thought} (CoT) prompting to medical reasoning for the task of automatic diagnosis.
Motivated by doctors' underlying reasoning process, we present \textit{Diagnostic-Reasoning CoT} (DR-CoT).
Empirical results demonstrate that by simply prompting large language models trained only on general text corpus with two DR-CoT exemplars, the diagnostic accuracy improves by 15\% comparing to standard prompting.
Moreover, the gap reaches a pronounced 18\% in out-domain settings.
Our findings suggest expert-knowledge reasoning in large language models can be elicited through proper promptings.
\end{abstract}

\section{Introduction}\label{sec:introduction}

\begin{wrapfigure}{r}{0.3\textwidth}
    \includegraphics[width=0.281\textwidth]{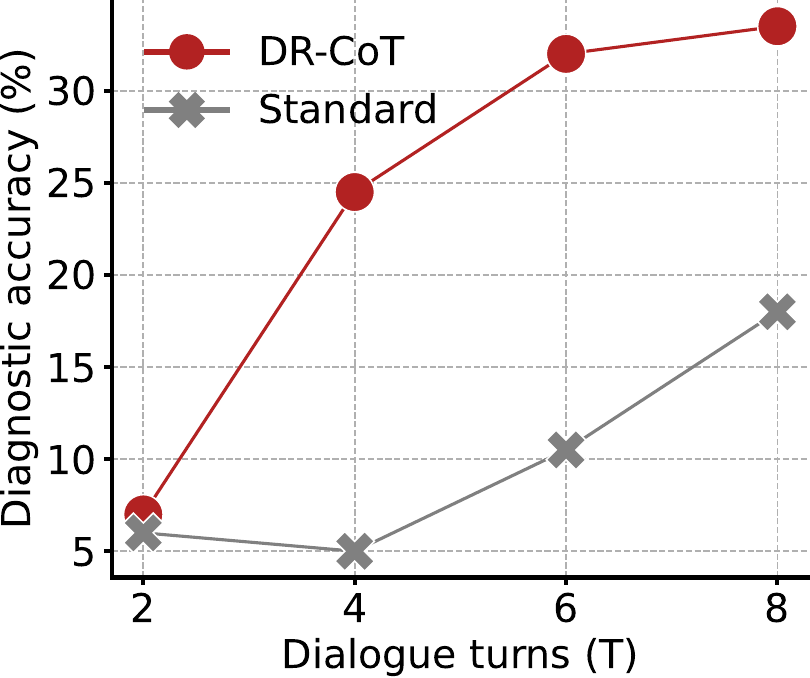}
    \setlength{\abovecaptionskip}{0pt}
    \setlength{\belowcaptionskip}{-10pt}
    \caption{DR-CoT achieves better diagnostic accuracy and converges more rapidly than standard prompting.}
    \label{fig:main-fig}
\end{wrapfigure}
Recently, \textit{Chain-of-Thought} (CoT) prompting~\citet{wei2022chain} has lunched a line of works on eliciting complex reasoning for large language models (LLMs)~\citep{kojima2022large,madaan-etal-2022-language,chen2022program,suzgun2022challenging}.
However, they mainly focus on arithmetic, commonsense, and symbolic reasoning tasks.
In this work, we extend the CoT framework to medical reasoning.
Specifically, we investigate the task of automatic diagnosis (AD), which requires knowledge-intensive, multi-step \textit{diagnostic reasoning}.

Typical patient-doctor interactions consist of an iterative question-answering (QA) sessions. 
In each dialogue turn, doctors inquire patients about clinical evidence (e.g., their backgrounds, symptoms experienced), and form a \textit{differential diagnosis} (DDx), a short list of possible diagnoses, in their underlying reasoning~\citep{graves2002users,rhoads2017formulating}.
The DDx then guides doctors in determining the next questions to ask, and patients' answers provide new evidence for refining the DDx.
Once the list is sufficiently narrowed, doctors can establish a final diagnosis.
We hypothesize such diagnostic reasoning process---from evidence to DDx to the next question---serves as the hidden CoT for AD.

An ideal dialogue system for automatic diagnosis (DSAD) requires the system to interact with patients in free-text natural language.
However, prior attempts mainly focus on the dialogue management component of DSAD, which only processes structured information (i.e., takes symptoms as input and output the next symptom to inquire or choose a diagnosis)~\citep{xia2020generative,liao2020task,ijcai2022p592}.
Moreover, most existing works use reinforcement learning, and can only predict 4$\sim$5 diagnoses due to the limitation of large action space for symptoms and the lack of holistic datasets~\citep{wei-etal-2018-task,xu2019end,zhao2021weighted}.

In this work, we propose:
\textbf{(1)} a few-shot LLM-based dialogue system for AD.
\textbf{(2)} a novel \textit{Diagnostic-Reasoning CoT} (DR-CoT) which elicits reasoning in LLMs towards better AD.
\textbf{(3)} a language-model-role-playing evaluation framework simulating realistic patient-doctor interactions.
To the best of our knowledge, we are the first to introduce LLMs into DSAD, and present a non-pipelined approach which in principle, has an unbounded diagnosis label set.
Empirical results show our dialogue system with DR-CoT prompting outperforms standard prompting by a striking 15\% on diagnostic accuracy~(Figure~\ref{fig:main-fig}), and the improvement holds for out-domain experiments~(Figure~\ref{fig:id-vs-od}).

\section{DR-CoT: Diagnostic-Reasoning Chain of Thought}
\noindent \textbf{Baseline.}
We construct our dialogue system based on InstructGPT~\citep{ouyang2022training}.
Inspired by Few-Shot Bot~\citep{madotto2021few}, we prompt the model with the template: ``\textit{\texttt{[I][S][D]}}", where \texttt{I}, \texttt{S}, and \texttt{D} denote the instruction, shots, and dialogue history, respectively (we provide an example in Appendix~\ref{appen:subsec:example-standard-prompt}).
Each shot in \texttt{S} is a \textit{complete} doctor-patient conversation, and \texttt{D} refers to the dialogue history of the current \textit{incomplete} conversation.
Concretely, given \texttt{D} $=\{(q_i, a_i)\}_{i=1,...,t-1}$, where $q_i$ is the utterance generated by our dialogue system, $a_i$ is the utterance from the patient, the output response at turn \textit{t} is $q_t$.
And for the next turn $t+1$, we prompt the model with the identical \texttt{I}, \texttt{S}, and an updated \texttt{D} $=\{(q_i, a_i)\}_{i=1,...,t}$.
Once the model deems the gathered evidence is sufficient, it generates a last utterance $d$ to establish the diagnosis and complete the dialogue.


\noindent \textbf{DR-CoT Prompting.}
To design the DR-CoT prompt, we simply apply two modifications comparing to the standard (i.e., baseline) prompt (we provide an example in Appendix~\ref{appen:subsec:example-DR-CoT-prompt}):
    \textbf{(1)} Augment \texttt{I} by instructing the model to first summarize the current observed evidence and formulate a DDx, then make inquiries based on the DDx.
    \textbf{(2)} Replace the $\{q_i\}_{i=2,...,T}\in$ \texttt{S} by ``\textcolor{BrickRed}{\textit{Based on the evidence\texttt{[E]}, the ranked differential diagnosis is\texttt{[DDx]}.
    To narrow down the differential diagnosis, the next question to ask is $q_i$.}}", where $T$ is the number of turns and \texttt{E} is the observed evidence in dialogue history.
The DDx guides the model to generate a crucial $q_i$ that leads to better evidence gathering, upon which the model compiles a more focused DDx, zooming in on the final diagnosis.

\begin{wrapfigure}{r}{0.3\textwidth}
    \includegraphics[width=0.3\textwidth]{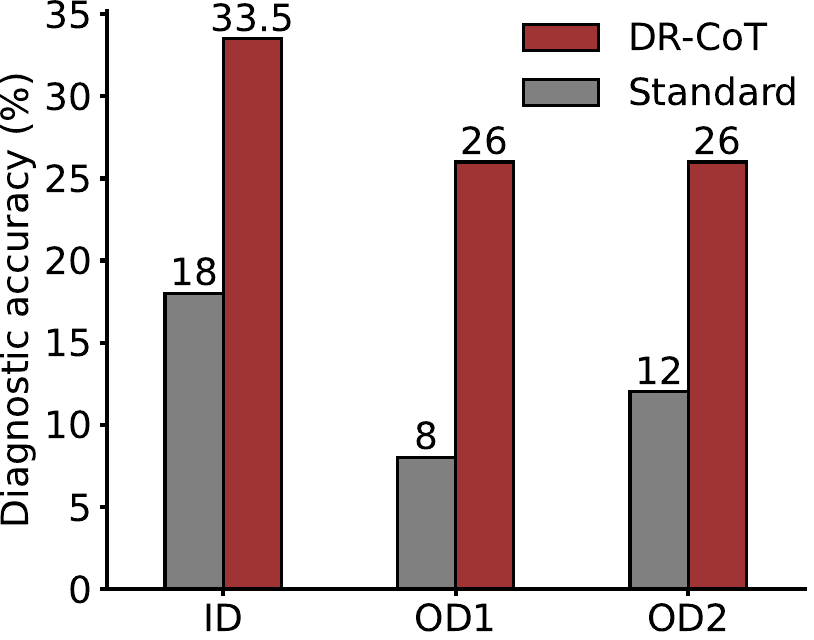}
    \setlength{\abovecaptionskip}{0pt}
    \setlength{\belowcaptionskip}{-10pt}
    \caption{The improvement on two out-domain testset suggests the elicited diagnostic reasoning ability is agnostic to specific exemplars.}
    \label{fig:id-vs-od}
\end{wrapfigure}

\noindent \textbf{Evaluation Framework.} Prior works generally evaluate DSAD by interacting with a lookup-table-like patient, which responds to inquiries in a binary, non-natural language fashion.
Such setting deviates from real clinical scenarios.
For a realistic testbed, we present a language-model-role-playing framework where the LLM plays both sides of the conversation, and performs evaluation by self-chat~\citep{li2019acute,ghandeharioun2019approximating}.

Specifically, after the DSAD generates a $q_i$ at turn $i$, we construct a \textit{patient bot} by prompting LLM with the template: "\textit{\texttt{[I][S][P][D]}}", where \texttt{P} is the patient profile (e.g., background information, symptoms, and medical history) and the last utterance of \texttt{D} is $q_i$ (we provide an example in Appendix~\ref{appen:subsec:example-patient-bot-prompt}).
The LLM is instructed to act as a patient and generate a response $a_i$ faithfully, namely, to answer $q_i$ based on the \textit{script} provided in \texttt{P}.
\footnote{We invite a physician to evaluate 100 sampled ($q_i,a_i$) pairs, where 95 $a_i$s are considered faithful w.r.t. \texttt{P}.}
We then append $a_i$ to \texttt{D} for prompting our DSAD.
In summary, the evaluation is fully automated by prompting LLMs to switch between a doctor role (i.e., the DSAD) and a patient role, and the interaction ends once the DSAD generates $d$.

\noindent \textbf{Results.} We adopt the DDXPlus~\citep{tchango2022ddxplus} dataset for our experiments.
Using two shots (i.e., exemplars), we conduct experiments with in-domain (ID) setting (the exemplars and test data share the same initial evidence; Figure~\ref{fig:main-fig}) and out-domain (OD) setting (with different initial evidence; Figure~\ref{fig:id-vs-od}).
Both settings exhibit substantial performance gain with our DR-CoT prompting.
To evaluate the accuracy in each dialogue turn $T$ (Figure~\ref{fig:main-fig}), if the final diagnosis has not been reached, we instruct the model to make the final diagnosis in turn $T$’s prompt (see in Appendix~\ref{appen:subsec:dialogue-turn} for details).
In addition, we conduct a small-scale human evaluation (see Appendix~\ref{appen:subsec:human-evalaution} for details and results) supporting our hypothesis that DR-CoT enables LLMs to ask more critical questions for establishing the correct diagnosis.
Details of other experimental setups including the DDXPlus dataset and the selection of in-domain and out-domain test sets are provided in~Appendix~\ref{appen:sec:experimental-setups}.

\section{Conclusion}
In this work, we propose DR-CoT prompting, which elicits the diagnostic reasoning ability in LLMs with striking empirical improvement, and introduce the first LLM-based DSAD along with a novel language-model-role-playing evaluation framework.
Our findings suggest through proper prompting, the potential expert-knowledge reasoning in LLMs can be unlocked for promising applicability.

\subsubsection*{URM Statement}
The authors acknowledge that at least one key author of this work meets the URM criteria of ICLR 2023 Tiny Papers Track.

\subsubsection*{Ethics and Reproducibility Statement}
In this work, we aim to provide a preliminary exploration into the diagnostic reasoning ability of large language models.
Note that the results are not validated in any real-world clinical settings.
Though with great potential of assisting clinicians in scenarios like telemedicine services (e.g., the doctors can review the interaction history and associated diagnosis predictions, before making final decisions or perform further inquiries), the current prototype of our dialogue system for automatic diagnosis is not yet directly applicable and requires thorough investigation.
To facilitate future research, the source code for conducting the experiments will be provided via a GitHub link in the camera-ready version upon acceptance.

\bibliography{iclr2023_conference_tinypaper}

\begin{thebibliography}{18}
\providecommand{\natexlab}[1]{#1}
\providecommand{\url}[1]{\texttt{#1}}
\expandafter\ifx\csname urlstyle\endcsname\relax
  \providecommand{\doi}[1]{doi: #1}\else
  \providecommand{\doi}{doi: \begingroup \urlstyle{rm}\Url}\fi

\bibitem[Chen et~al.(2022)Chen, Ma, Wang, and Cohen]{chen2022program}
Wenhu Chen, Xueguang Ma, Xinyi Wang, and William~W Cohen.
\newblock Program of thoughts prompting: Disentangling computation from
  reasoning for numerical reasoning tasks.
\newblock \emph{arXiv preprint arXiv:2211.12588}, 2022.

\bibitem[Ghandeharioun et~al.(2019)Ghandeharioun, Shen, Jaques, Ferguson,
  Jones, Lapedriza, and Picard]{ghandeharioun2019approximating}
Asma Ghandeharioun, Judy~Hanwen Shen, Natasha Jaques, Craig Ferguson, Noah
  Jones, Agata Lapedriza, and Rosalind Picard.
\newblock Approximating interactive human evaluation with self-play for
  open-domain dialog systems.
\newblock \emph{Advances in Neural Information Processing Systems}, 32, 2019.

\bibitem[Graves(2002)]{graves2002users}
Rebecca~S Graves.
\newblock Users' guides to the medical literature: A manual for evidence-based
  clinical practice.
\newblock \emph{Journal of the Medical Library Association}, 90\penalty0
  (4):\penalty0 483, 2002.

\bibitem[Kojima et~al.(2022)Kojima, Gu, Reid, Matsuo, and
  Iwasawa]{kojima2022large}
Takeshi Kojima, Shixiang~Shane Gu, Machel Reid, Yutaka Matsuo, and Yusuke
  Iwasawa.
\newblock Large language models are zero-shot reasoners.
\newblock \emph{arXiv preprint arXiv:2205.11916}, 2022.

\bibitem[Li et~al.(2019)Li, Weston, and Roller]{li2019acute}
Margaret Li, Jason Weston, and Stephen Roller.
\newblock Acute-eval: Improved dialogue evaluation with optimized questions and
  multi-turn comparisons.
\newblock \emph{arXiv preprint arXiv:1909.03087}, 2019.

\bibitem[Liao et~al.(2020)Liao, Liu, Wei, Peng, Chen, Sun, and
  Huang]{liao2020task}
Kangenbei Liao, Qianlong Liu, Zhongyu Wei, Baolin Peng, Qin Chen, Weijian Sun,
  and Xuanjing Huang.
\newblock Task-oriented dialogue system for automatic disease diagnosis via
  hierarchical reinforcement learning.
\newblock \emph{arXiv preprint arXiv:2004.14254}, 2020.

\bibitem[Liu et~al.(2022)Liu, Cheng, Wang, Tang, Liu, Zhao, Li, Zheng, and
  Liang]{ijcai2022p592}
Wenge Liu, Yi~Cheng, Hao Wang, Jianheng Tang, Yafei Liu, Ruihui Zhao, Wenjie
  Li, Yefeng Zheng, and Xiaodan Liang.
\newblock “my nose is running.” “are you also coughing?”: Building a
  medical diagnosis agent with interpretable inquiry logics.
\newblock In Lud~De Raedt (ed.), \emph{Proceedings of the Thirty-First
  International Joint Conference on Artificial Intelligence, {IJCAI-22}}, pp.\
  4266--4272. International Joint Conferences on Artificial Intelligence
  Organization, 7 2022.
\newblock \doi{10.24963/ijcai.2022/592}.
\newblock URL \url{https://doi.org/10.24963/ijcai.2022/592}.
\newblock Main Track.

\bibitem[Madaan et~al.(2022)Madaan, Zhou, Alon, Yang, and
  Neubig]{madaan-etal-2022-language}
Aman Madaan, Shuyan Zhou, Uri Alon, Yiming Yang, and Graham Neubig.
\newblock Language models of code are few-shot commonsense learners.
\newblock In \emph{Proceedings of the 2022 Conference on Empirical Methods in
  Natural Language Processing}, pp.\  1384--1403, Abu Dhabi, United Arab
  Emirates, December 2022. Association for Computational Linguistics.
\newblock URL \url{https://aclanthology.org/2022.emnlp-main.90}.

\bibitem[Madotto et~al.(2021)Madotto, Lin, Winata, and Fung]{madotto2021few}
Andrea Madotto, Zhaojiang Lin, Genta~Indra Winata, and Pascale Fung.
\newblock Few-shot bot: Prompt-based learning for dialogue systems.
\newblock \emph{arXiv preprint arXiv:2110.08118}, 2021.

\bibitem[Ouyang et~al.(2022)Ouyang, Wu, Jiang, Almeida, Wainwright, Mishkin,
  Zhang, Agarwal, Slama, Ray, et~al.]{ouyang2022training}
Long Ouyang, Jeff Wu, Xu~Jiang, Diogo Almeida, Carroll~L Wainwright, Pamela
  Mishkin, Chong Zhang, Sandhini Agarwal, Katarina Slama, Alex Ray, et~al.
\newblock Training language models to follow instructions with human feedback.
\newblock \emph{arXiv preprint arXiv:2203.02155}, 2022.

\bibitem[Rhoads et~al.(2017)Rhoads, Penick, et~al.]{rhoads2017formulating}
Jacqueline Rhoads, Julie~C Penick, et~al.
\newblock \emph{Formulating a Differential Diagnosis for the Advanced Practice
  Provider}.
\newblock Springer Publishing Company, 2017.

\bibitem[Suzgun et~al.(2022)Suzgun, Scales, Sch{\"a}rli, Gehrmann, Tay, Chung,
  Chowdhery, Le, Chi, Zhou, et~al.]{suzgun2022challenging}
Mirac Suzgun, Nathan Scales, Nathanael Sch{\"a}rli, Sebastian Gehrmann, Yi~Tay,
  Hyung~Won Chung, Aakanksha Chowdhery, Quoc~V Le, Ed~H Chi, Denny Zhou, et~al.
\newblock Challenging big-bench tasks and whether chain-of-thought can solve
  them.
\newblock \emph{arXiv preprint arXiv:2210.09261}, 2022.

\bibitem[Tchango et~al.(2022)Tchango, Goel, Wen, Martel, and
  Ghosn]{tchango2022ddxplus}
Arsene~Fansi Tchango, Rishab Goel, Zhi Wen, Julien Martel, and Joumana Ghosn.
\newblock Ddxplus: A new dataset for automatic medical diagnosis.
\newblock \emph{Proceedings of the Neural Information Processing Systems-Track
  on Datasets and Benchmarks}, 2, 2022.

\bibitem[Wei et~al.(2022)Wei, Wang, Schuurmans, Bosma, Chi, Le, and
  Zhou]{wei2022chain}
Jason Wei, Xuezhi Wang, Dale Schuurmans, Maarten Bosma, Ed~Chi, Quoc Le, and
  Denny Zhou.
\newblock Chain of thought prompting elicits reasoning in large language
  models.
\newblock \emph{arXiv preprint arXiv:2201.11903}, 2022.

\bibitem[Wei et~al.(2018)Wei, Liu, Peng, Tou, Chen, Huang, Wong, and
  Dai]{wei-etal-2018-task}
Zhongyu Wei, Qianlong Liu, Baolin Peng, Huaixiao Tou, Ting Chen, Xuanjing
  Huang, Kam-fai Wong, and Xiangying Dai.
\newblock Task-oriented dialogue system for automatic diagnosis.
\newblock In \emph{Proceedings of the 56th Annual Meeting of the Association
  for Computational Linguistics (Volume 2: Short Papers)}, pp.\  201--207,
  Melbourne, Australia, July 2018. Association for Computational Linguistics.
\newblock \doi{10.18653/v1/P18-2033}.
\newblock URL \url{https://aclanthology.org/P18-2033}.

\bibitem[Xia et~al.(2020)Xia, Zhou, Shi, Lu, and Huang]{xia2020generative}
Yuan Xia, Jingbo Zhou, Zhenhui Shi, Chao Lu, and Haifeng Huang.
\newblock Generative adversarial regularized mutual information policy gradient
  framework for automatic diagnosis.
\newblock In \emph{Proceedings of the AAAI conference on artificial
  intelligence}, pp.\  1062--1069, 2020.

\bibitem[Xu et~al.(2019)Xu, Zhou, Gong, Liang, Tang, and Lin]{xu2019end}
Lin Xu, Qixian Zhou, Ke~Gong, Xiaodan Liang, Jianheng Tang, and Liang Lin.
\newblock End-to-end knowledge-routed relational dialogue system for automatic
  diagnosis.
\newblock In \emph{Proceedings of the AAAI conference on artificial
  intelligence}, pp.\  7346--7353, 2019.

\bibitem[Zhao et~al.(2021)Zhao, Chen, and Chen]{zhao2021weighted}
Xinyan Zhao, Liangwei Chen, and Huanhuan Chen.
\newblock A weighted heterogeneous graph-based dialog system.
\newblock \emph{IEEE Transactions on Neural Networks and Learning Systems},
  2021.

\end{thebibliography}
\bibliographystyle{iclr2023_conference_tinypaper}

\newpage
\section*{\centering Outline of Appendix}
We provide additional details in this appendix for our work entitled ``\textit{Large Language Models Perform Diagnostic Reasoning}".
The content is organized as follows:
\begin{itemize}
    \item Section~\ref{appen:sec:example-prompts} contains examples of standard prompting (i.e., baseline) (Section~\ref{appen:subsec:example-standard-prompt}), DR-CoT prompting (Section~\ref{appen:subsec:example-DR-CoT-prompt}), and patient bot prompting (Section~\ref{appen:subsec:example-patient-bot-prompt}).
    \item Section~\ref{appen:sec:experimental-setups} contains details of DDXPlus and the selection for in-domain and out-domain samples (Section~\ref{appen:subsec:dataset}), the model (Section~\ref{appen:subsec:model}), and the dialogue turn setting (Section~\ref{appen:subsec:dialogue-turn}).
    \item Section~\ref{appen:subsec:human-evalaution} contains human evaluation results for qualitative assessment of DR-CoT.
\end{itemize}
\appendix
\section{Example Prompts}\label{appen:sec:example-prompts}
\subsection{An Example of Standard Prompting (Baseline)}\label{appen:subsec:example-standard-prompt}
\begin{figure}[h]
    \centering
    \includegraphics[width=0.8\textwidth]{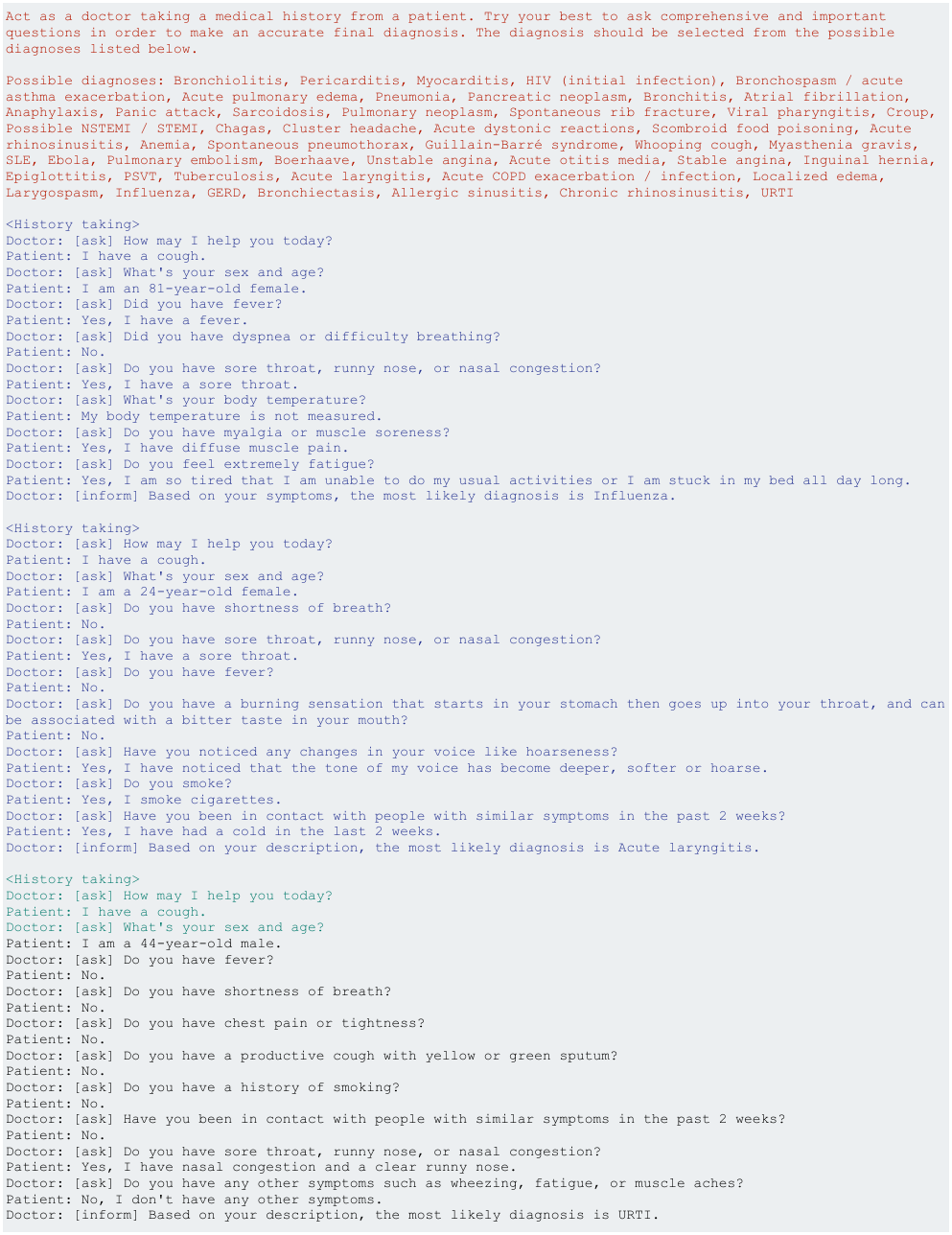}
    \caption{The initial prompt includes \textcolor{BrickRed}{the instruction \texttt{I}}, \textcolor{BlueViolet}{the shots \texttt{S}}, and \textcolor{JungleGreen}{the input \texttt{D}}.
    The generated question $q_i$ of the prompted model (i.e., the DSAD) and the answer $a_i$ from the patient bot is presented in the remaining text in black.
    }
\end{figure}

\newpage
\subsection{An Example of DR-CoT prompting}\label{appen:subsec:example-DR-CoT-prompt}
\begin{figure}[h]
    \centering
     \includegraphics[width=0.8\textwidth]{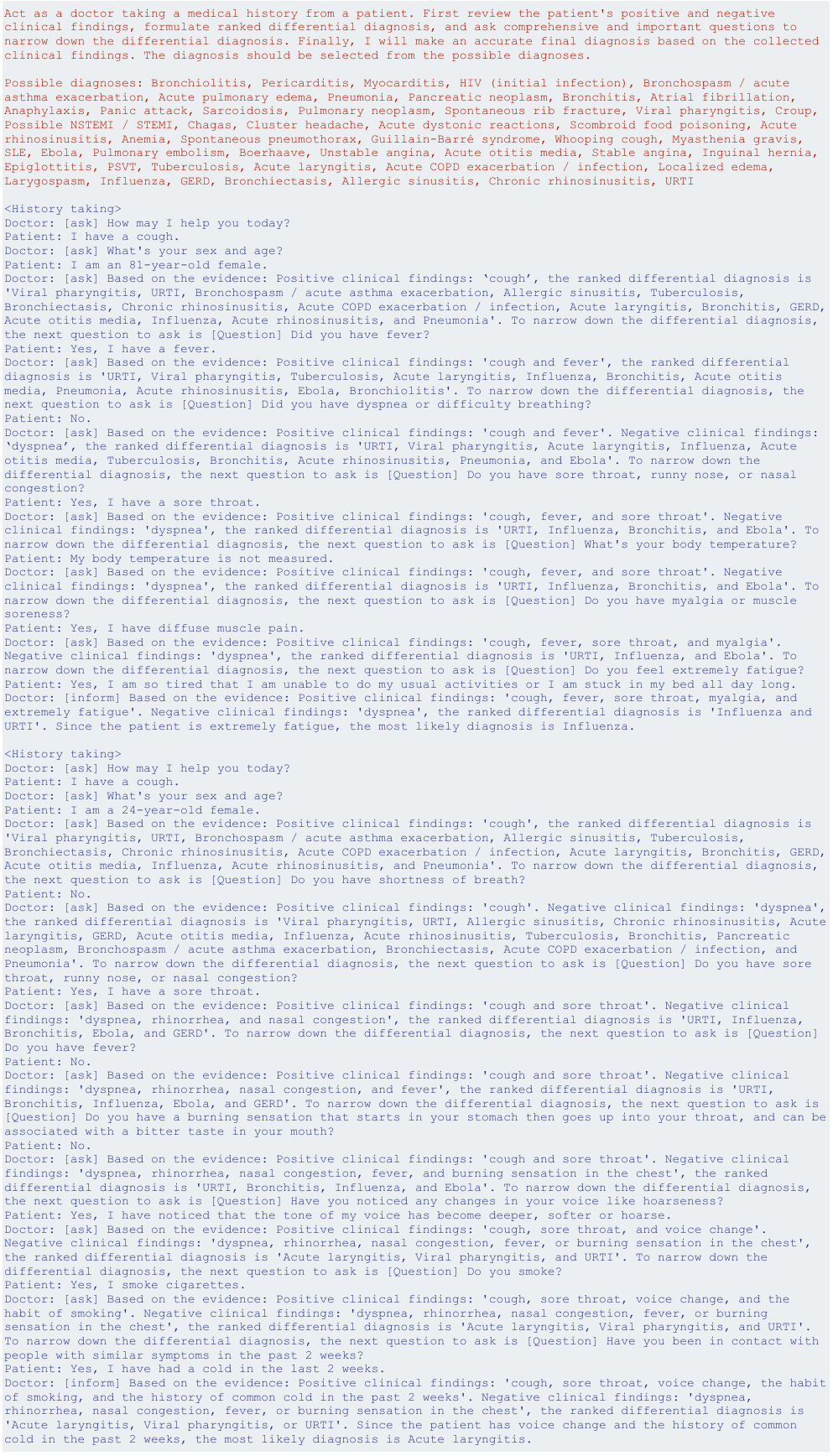}
\end{figure}
\newpage
\begin{figure}[h]
    \centering
    \includegraphics[width=0.8\textwidth]{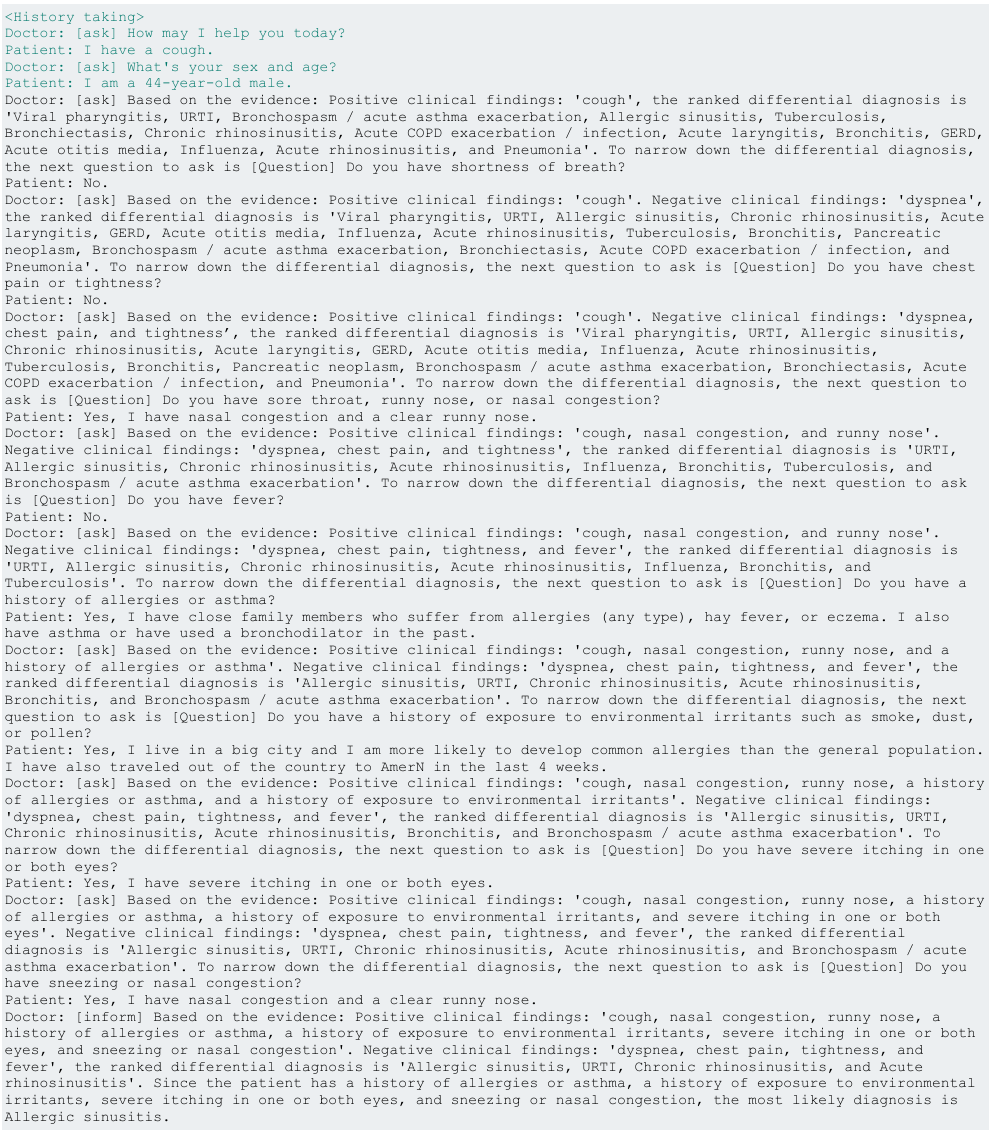}
    \caption{The initial prompt includes \textcolor{BrickRed}{the instruction \texttt{I}}, \textcolor{BlueViolet}{the shots \texttt{S}}, and \textcolor{JungleGreen}{the input \texttt{D}}.
    The generated question $q_i$ of the prompted model (i.e., the DSAD) and the answer $a_i$ from the patient bot is presented in the remaining text in black.
    We invite a physician to write the \texttt{DDx} part of the two DR-CoT exemplars.
    }
\end{figure}

\newpage
\subsection{An Example of Patient Bot Prompting}\label{appen:subsec:example-patient-bot-prompt}
\begin{figure}[h]
    \centering
    \includegraphics[width=0.8\textwidth]{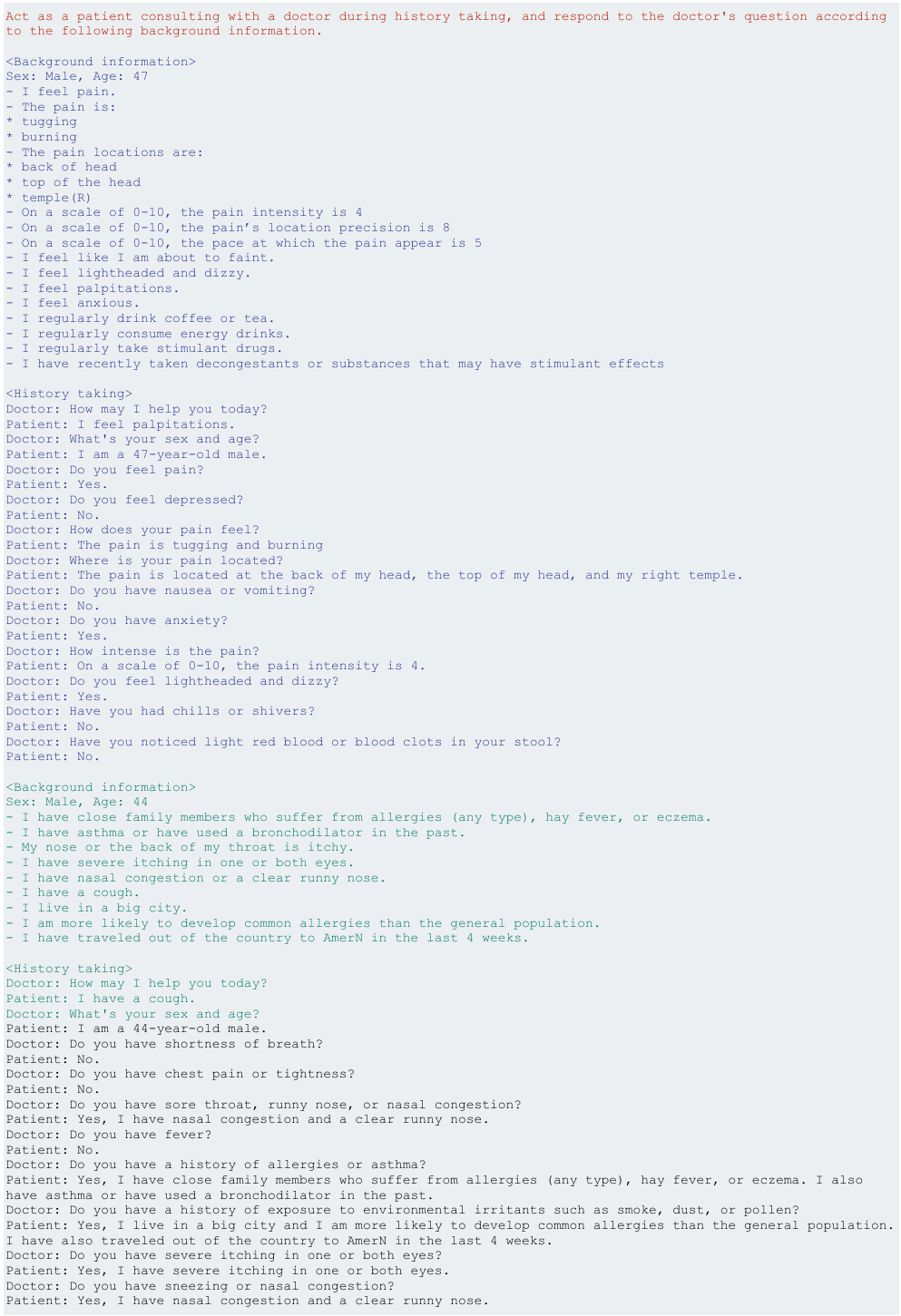}
    \caption{The initial prompt includes \textcolor{BrickRed}{the instruction \texttt{I}}, \textcolor{BlueViolet}{the shot \texttt{S}}, and \textcolor{JungleGreen}{the input \texttt{P} and \texttt{D}}.
    The generated answer $a_i$ of the prompted model (i.e., the patient bot) and the inquired question $q_i$ from the DSAD is presented in the remaining text in black.
    }
\end{figure}

\newpage
\section{Details of Experimental setups}\label{appen:sec:experimental-setups}
\subsection{Dataset}\label{appen:subsec:dataset}
\noindent \textbf{Details of DDXPlus.} Compared with other AD datasets with only 4$\sim$5 diagnoses and 40$\sim$70 types of evidence~\citep{wei-etal-2018-task,xu2019end}, the DDXPlus dataset includes 49 diagnoses and 223 types of evidence.
Each instance of DDXPlus~\citep{tchango2022ddxplus} represents a \textit{patient}, and we use the following attributes of the patient for our experiments, : \texttt{AGE}, \texttt{SEX}, \texttt{INITIAL\_EVIDENCE} (\texttt{IE}), \texttt{EVIDENCES} (i.e., the symptoms or other clinical findings experienced by the patient), and \texttt{PATHOLOGY}.
The \texttt{AGE}, \texttt{SEX}, \texttt{IE}, and \texttt{EVIDENCES} are used to construct the patient profile \texttt{P} based on a rule-based template, and the \texttt{PATHOLOGY} is the ground truth diagnosis for computing the diagnostic accuracy.
At the beginning of each dialogue session, \texttt{AGE}, \texttt{SEX}, and \texttt{IE} are provided to the doctor model to kick-start the dialogue.

\noindent \textbf{In-domain and Out-domain.} We refer to the test samples that share the same \texttt{IE} with our two exemplars as \textit{in-domain}, and the ones that do not as \textit{out-domain}.
The rationale behind this notion is that the LLM only sees in-domain exemplars' \texttt{IE}, thus, other \texttt{IE}s are considered to be out-domain. 
For the selection of in-domain (ID) and out-domain (OD) \texttt{IE}, we choose \textit{cough}, \textit{dyspnea}, and \textit{runny nose} (i.e., \textit{toux}, \textit{dyspn}, and \textit{rhino\_clair} in DDXPlus) as the \texttt{IE}s for ID, OD1, and OD2 respectively.
These three symptoms is the most frequently occurred \texttt{IE} in the testset of DDXPlus with 618, 11305, and 6270 instances, respectively.
As accessing LLMs is costly, especially with our limited budget, we randomly sample 200, 100, and 100 instances for ID, OD1, and OD2 to conduct our~experiments.

\subsection{Model}\label{appen:subsec:model}
For the model choice, we adopt the \texttt{text-davinci-003} version of the InstructGPT via OpenAI API\footnote{https://platform.openai.com/docs/models/gpt-3}.
To the best of our knowledge, it is the most capable version of publicly available GPT-3 at the time we conduct our experiments.
As for hyperparameters, we set the maximum tokens to 384 and the temperature to 0 (since diversity is not of necessity for our task).
The remaining hyperparameters are kept as the default values.

\subsection{Dialogue Turn}\label{appen:subsec:dialogue-turn}
To evaluate the diagnosis accuracy of dialogue turn $T$, if the final diagnosis has not been generated, we instruct the DSAD to make the final diagnosis by inserting a prefix ``\texttt{[inform]}" in turn $T$’s prompt (see Appendix~\ref{appen:subsec:example-standard-prompt}, \ref{appen:subsec:example-DR-CoT-prompt}).
And to avoid lengthy dialogue, we set a pre-defined $T_{max}$ as the maximum number of turn allowed for making the final diagnosis.
We determine $T_{max}$ by inviting a physician to interact with our patient bot, and find that 6$\sim$8 turns are required on average.
Thus, we set $T_{max}=8$ for our experiments.
Similarly, if the DSAD fails to give a diagnosis before the 8-$th$ turn, the prefix ``\texttt{[inform]}" will be given to the DSAD to force the prediction.

\section{Human Evaluation}\label{appen:subsec:human-evalaution}
One of our hypotheses is that DR-CoT guides LLMs to generate a better $q_i$ for collecting crucial evidence, which ultimately leads to a more accurate final diagnosis.
Establishing an accurate final diagnosis depends on two abilities: (1) asking critical questions to collect evidence, and (2) predicting the diagnosis based on the collected evidence.
To verify that DR-CoT prompted LLMs indeed ask better questions, we invite a physician to evaluate the dialogues as there is no ground truth for~$q_i$.
During the evaluation, only the $(q_i,a_i)$ pairs are revealed to anonymize the promptimg methods.
Table~\ref{tab:human-eval} summarizes the results, which show that DR-CoT prompting enables LLMs to ask questions which are more critical for establishing the correct diagnosis.
\begin{table}[h]
  \caption{Number of dialogues where the $q_i$s are deemed more critical for diagnosing the patient.}
  \label{tab:human-eval}
  \begin{center}
  \small
    \begin{tabular}{lrrrr}
    \toprule
    \textbf{Method} & \multicolumn{1}{l}{\textbf{ID}} & \multicolumn{1}{l}{\textbf{OD1}} & \multicolumn{1}{l}{\textbf{OD2}} \\
    \midrule
    Standard  & 2   & 2  & 1 \\
    DR-CoT    & 18  & 8  & 9 \\
    \midrule
    Total     & 20  & 10 & 10 \\
    \bottomrule
    \end{tabular}%
  \end{center}
\end{table}

\end{document}